\documentclass{article}

\usepackage[final,nonatbib]{NeuRIPS2019/neurips_2019}

\usepackage[utf8]{inputenc} 
\usepackage[T1]{fontenc}    
\usepackage{hyperref}       
\usepackage{url}            
\usepackage{booktabs}       
\usepackage{amsfonts}       
\usepackage{nicefrac}       
\usepackage{microtype}      

\usepackage{amsmath}
\DeclareMathOperator*{\argmax}{arg\,max}

\usepackage{algorithm,algorithmic}
\usepackage{multicol}
\usepackage{changepage}
\usepackage[skins]{tcolorbox}
\usepackage{dsfont}
\usepackage{subcaption}
\usepackage{sidecap}
\usepackage{wrapfig}
\usepackage{xr-hyper}
\usepackage{xcolor}
\definecolor{ceruleanblue}{rgb}{0.16, 0.32, 0.75}
\definecolor{darkmagenta}{rgb}{0.55, 0.0, 0.55}
\definecolor{fluorescentorange}{rgb}{1.0, 0.75, 0.0}
\definecolor{charcoal}{rgb}{0.21, 0.27, 0.31}

\usepackage[title]{appendix}

\allowdisplaybreaks

\makeatletter
\newcommand*{\addFileDependency}[1]{
  \typeout{(#1)}
  \@addtofilelist{#1}
  \IfFileExists{#1}{}{\typeout{No file #1.}}
}
\makeatother
 
\newcommand*{\myexternaldocument}[1]{%
    \externaldocument{#1}%
    \addFileDependency{#1.tex}%
    \addFileDependency{#1.aux}%
}

\myexternaldocument{supplementary}

\title{Emergence of Pragmatics from Referential Game between Theory of Mind Agents}

\author{%
  Luyao Yuan\\
  \texttt{yuanluyao@ucla.edu} \\
   \And
   Zipeng Fu \\
   \texttt{fu-zipeng@engineering.ucla.edu}\\
   \And
   Jingyue Shen \\
   \texttt{brianshen@ucla.edu}\\
   \And
   Lu Xu \\
   \texttt{luxunxn7@ucla.edu} \\
   \And
   Juhong Shen \\
   \texttt{jhshen@ucla.edu} \\
   \And
   Song-Chun Zhu \\
   \texttt{sczhu@cs.ucla.edu}\\
   \AND
    \normalfont{Department of Computer Science}\\
    University of California, Los Angeles \\
}

\begin{document}

\maketitle

\begin{abstract}
Pragmatics studies how context can contribute to language meanings. In human communication, language is never interpreted out of context, and sentences can usually convey more information than their literal meanings. However, this mechanism is missing in most multi-agent systems, restricting the communication efficiency and the capability of human-agent interaction. In this paper, we propose an algorithm, using which agents can spontaneously learn the ability to “read between lines” without any explicit hand-designed rules. We integrate theory of mind (ToM) in a cooperative multi-agent pedagogical situation and propose an adaptive reinforcement learning (RL) algorithm to develop a communication protocol. ToM is a profound cognitive science concept, claiming that people regularly reason about other's mental states, including beliefs, goals, and intentions, to obtain performance advantage in competition, cooperation or coalition. With this ability, agents consider language as not only messages but also rational acts reflecting others hidden states. Our experiments demonstrate the advantage of pragmatic protocols over non-pragmatic protocols. We also show the teaching complexity following the pragmatic protocol empirically approximates to recursive teaching dimension (RTD).
\end{abstract}

\section{Introduction}
The study of emergent languages has become an important topic in cognitive science and artificial intelligence for years, as effective communication is the prerequisite for successful cooperation in both human society~\cite{christiansen2003language,ibsen2018language} and multi-agent systems (MAS)~\cite{goldman2007learning,foerster2016learning,lazaridou2018emergence}. Communication, as Grice characterized in his original model of pragmatics~\cite{grice1975logic}, should follow cooperative principles, where listeners make exquisitely sensitive inferences about what utterances mean given their knowledge of the speaker, the language, and the context, and the speaker collaboratively makes contributions to the conversational goals~\cite{vogel2013emergence,goodman2016pragmatic}. For example, if there are three boys in a class and teacher $A$ told teacher $B$ that “\textbf{some} of the boys went to the party”, then teacher $B$ will legitimately infer that one or two of the boys went to the party~\cite{grice1975logic,jager2012game}. Although according to the literal meaning, it could be that all three of the boys went to the party, teacher $B$ is less likely to interpret the utterance that way, because teacher $A$, a cooperative speaker, would have said “\textbf{all} of the boys went to the party” in that case to prevent ambiguity. This seemingly trivial example (called scalar implicature in pragmatics) illustrates that pragmatic conversation is, in most of the time, taken for granted in human communication, and shows how significant hidden information can be acquired from literal meanings. However, current MAS tend to model communication merely as information exchange between agents, among which messages are deciphered only by their literal meanings~\cite{kinney1998learning,bernstein2002complexity,goldman2007learning,sukhbaatar2016learning}. Even with perfect understanding among each other, this type of communication cannot achieve optimal efficiency, as the intention of the communicator implicitly suggested by the message is ignored. 

To perform pragmatic communication, the listener needs to not only comprehend the speaker's utterance, but also infer his mental states. It has been shown that humans, during an interaction, can reason about others’ beliefs, goals, intentions and predict opponent/partner's behaviors~\cite{premack1978does,yoshida2008game,baker2017rational}, a capability called ToM. In some cases, people can even use ToM recursively, and form beliefs about the way others reason about themselves~\cite{de2015higher}. Thus, in order to collaborate and communicate with people smoothly, artificial agents must also bear similar potentially recursive mutual reasoning capability. Despite the recent surge of multi-agent collaboration modeling~\cite{kinney1998learning,sukhbaatar2016learning,das2017learning,foerster2018counterfactual}, integrating ToM is still a nontrivial challenge. A few approaches attempted to model nested belief of other agents in general multi-agent systems, but extensive computation restricts the scale of the solvable problems~\cite{doshi2009monte,han2018learning}. When an agent has an incomplete observation of the environment, it needs to form a belief, a distribution over the actual state of the environment, to take actions~\cite{yoshida2008game,han2018learning}. ToM agents, besides their own beliefs about the state, or 0-th level beliefs, also model other agents' beliefs, forming 1-st level beliefs. They can further have beliefs about others' 1-st level beliefs about their 0-th level belief, so on and so forth~\cite{doshi2009monte,de2014theory,yoshida2008game,de2015higher,de2017estimating}. The intractability of distribution over distribution makes exact solving for ToM agents' nested beliefs extremely complicated~\cite{doshi2009monte}.

Therefore, an approach to acquire the sophistication of high-level recursions without getting entangled into the curse of intractability is needed. In this paper, we propose an adaptive training process, following which pragmatic communication protocol can emerge between cooperative ToM agents modeling only the 1-st level belief over belief. The complexity of higher level recursions can be preserved by the dynamic evolving of agents' tractable belief estimation functions. We don't assume agent has a certain level of recursions, which requires modeling nested beliefs from the 0-th level up to the desired level~\cite{doshi2009monte,de2014theory,de2015higher,de2017estimating}. Instead, we directly learn a function to approximate partners' actual beliefs and how to react accordingly. In cooperative games, this learning becomes mutual adaptation, with controlled exploration rate, improving the performance of the multi-agent system~\cite{claus1998dynamics}. Intuitively, for a pair of agents, we update them alternatively, namely, fixing one while training the other, simulating the iterative best response (IBR) model, which proves to converge to a fixed point Nash equilibrium in strong and weak interpretation game~\cite{jager2012game}. We demonstrate the effectiveness and advantage of the pragmatic protocol with referential games, a communication game widely used in linguistic and cognitive studies in the context of language evolution~\cite{lazaridou2016multi,cao2018emergent,lazaridou2018emergence}. It provides a good playground for pragmatic and pedagogical interactions between a teacher and a student and can easily generalize to a comprehensive teaching task with a large concept space. 

We tested our algorithm and evaluated the pragmatic teaching protocol with both symbolic and pixel data, and achieved significant performance gain over previous algorithms in terms of referential accuracy. Also, we found that if messages are grounded with human expertise before interaction, the emerged protocol achieves teaching complexity empirically approximates recursive teaching dimension (RTD)~\cite{doliwa2014recursive}, the worst-case number of examples required for cooperative agents concept learning~\cite{chen2016recursive}.

This paper makes two major contributions: 1) we showed that a pragmatic communication protocol can emerge between ToM agents through adaptive reinforcement learning. This protocol significantly improves the cooperative multi-agent communication by enabling the agents to extract hidden meanings from the context to enrich the literal information; 2) we proposed an algorithm to develop protocols empirically approximating the teaching complexity bound between cooperative agents provided by RTD. Extensive experiments on both the symbolic and 3D object datasets demonstrate the effectiveness of our proposed protocol.


\section{Background: Referential Game}
There are a teacher and a student in a referential game. The teacher has a target in mind and aims to send a message to the student so that the student can identify the target out of a set of distractors after receiving this message. The motivation of our algorithm is the rational speech act (RSA) model between ToM agents (also termed as bilateral optimality~\cite{blutner2000some}): in order to establish proper communication, the speaker has to take into account the perspective of the listener, while the listener has to take into account the perspective of the teacher~\cite{goodman2016pragmatic}. Figure \ref{fig:game} shows an example. There are three objects, a blue sphere, a red sphere, and a blue cone. Suppose the target is the blue sphere. If the only allowed messages are colors and shapes, then, for a literal student, there is no unique identifier for the blue sphere, because both “blue” and “sphere” have more than one consistent candidates. Nonetheless, a pragmatic student, after hearing “blue” from the teacher, should be able to do counterfactual reasoning and identify the blue sphere instead of the blue cone, because he knows the teacher is helpful and would have used cone to refer to the blue cone unambiguously. During pragmatic communication, a message conveys more information than the message itself. The usage of that message can usually suggest the intention of the teacher.

The referential game can be formally defined by a tuple $\langle A, B, \Omega,$  $\mathcal{M}, \mathcal{A}\rangle$, where $A$ and $B$ stand for a teacher and a student. $\Omega$ is the instance space, where the distractors and targets are sampled from. $\mathcal{M}$ is the message space and $\mathcal{A}$ is the student's action space. In a specific game, a set of instances $O \subseteq \Omega$ is sampled from $\Omega$ as candidates, and one of the candidates $o^{\star} \in O$ is designated as the target, while the rest, $O/\{o^{\star}\}$, are distractors. The candidates $O$ are available to both of the agents, while only the teacher knows the target, $o^{\star}$. Agents take turns in this game. In every round, the teacher first sends a message $m_t \in \mathcal{M}$ to the student, followed by an action $a_t \in \mathcal{A} = \{1, 2, ..., |O|, \Xi\}$ taken by the student, where number 1 to $|O|$ represent "identify a certain instance as the target", $\Xi$ means "wait for next message" and $t$ stamps the $t$-th round. Every message comes with a message cost, $c_m$, and the total gain for both the teacher and the student, given $a_T$ the first non $\Xi$ action, is $R = \sum_{t = 1}^T -c_{m_t} + \mathbf{1}(o^{\star} = O[a_T])$. Notice that the game ends when the student performs a non $\Xi$ action. We define a protocol between $A$ and $B$ as a set of policies 
\begin{align*}
    \Pi = \langle &\pi_A: \mathbb{P}(\Omega) \times O \times \mathcal{M^*} \times \mathcal{M} \rightarrow [0, 1] , \pi_B : \mathbb{P}(\Omega) \times \mathcal{M^*} \times \mathcal{M} \times \mathcal{A} \rightarrow [0, 1] \rangle
\end{align*}
$\mathbb{P}(\Omega)$ is the power set of $\Omega$, where $O$ sampled from, and $*$ is the kleene star, standing for the history of message. Intuitively, the teacher selects a message based on the distractors, the target and communication history. The student chooses an action according to candidates, history and the latest message. The goal for both of the agents is to maximize the expected gain:
\begin{align}
\label{eq:gain}
    \mathbf{E}_{\substack{O \sim \mathbb{P}(\Omega), o^{\star} \sim O,\\m_{1:T} \sim \pi_A, a_{1:T} \sim \pi_B}}\left[-\sum_{t = 1}^T c_{m_t} + \mathbf{1}(o^{\star} = O[a_T])\right]
\end{align}
\section{Related Work}
\textbf{Pragmatics}~\cite{grice1975logic} has profound linguistic and cognitive science origin dating back to 1970s. However, the integration of this topic with multi-agent games only draws people's attention in this decade. The RSA model was raised by Golland et al.~\cite{golland2010game} and developed in~\cite{frank2012predicting,shafto2014rational,goodman2016pragmatic,andreas2016reasoning} in later works. Yet, all of these works model only single utterance and require hand designed production rules for the agents, while in our algorithm, all policies are learned by the agents and multi-round of communication is allowed. \cite{frank2009informative,vogel2013emergence,khani2018planning} model pragmatic reasoning for human action prediction, but they all require domain-specific engineering, and pragmatically annotated training data. In our work, all models are trained by self-playing with RL. \cite{vogel2013emergence} needs human data to initialize the literal speaker and uses RL to train listener with pragmatic reasoning, but they assumes fixed literal speaker. On the contrary, we endow both the teacher and the student ToM and allow them to do mutual pragmatic reasoning. The discriminative best response algorithm in \cite{vogel2014learning} is also inspired the IBR model~\cite{jager2012game}, but they used supervised learning given language data instead of RL. Also, our game states are more complicated than their toy examples.

\textbf{Emergence of language in communication games}. Emergence of language is a topic about a group of agents developing a communication protocol to complete a task, which can either be cooperative or competitive. In most recent studies, agents start with a set of ungrounded symbols and first learn to ground these symbols using reinforcement learning approaches supervised by the task rewards. The major novelty of our work comparing with previous methods is that pragmatic reasoning can emerge from our protocols without explicit rules coded. 

In~\cite{lazaridou2018emergence}, two agents try to develop a protocol through playing the referential game, in which the teacher sees only the target but no distractors, eliminating the possibility of taking advantage of ToM, as no counterfactual reasoning can happen at the student side. In~\cite{lazaridou2016multi} the teacher sends a message according to the context of candidates, but no student reaction is simulated before the selection. In~\cite{bogin2018emergence,choi2018compositional}, the teacher can simulate the reaction of a fixed student, who does not  model teacher's mind and cannot benefit from counterfactual reasoning about teacher's intention behind messages, terminating the recursive process after only one cycle.

A variation of the referential game was played in~\cite{evtimova2017emergent}, where the teacher sees an image and the student sees a set of descriptions. The goal is for the student to identify a suitable description for the teacher's image by multi-round communication. A similar multi-round communication game using natural language was also played in~\cite{das2017learning}. In these games, the student can ask further questions after receiving the first message and is in charge of the final decision making, while the teacher will answer the question based on the target and the communication history. These works focus on learning shared embedding between objects and messages instead of grounding messages to attributes that reappear across the object space. Also, neither of these papers include agents who model their partner's minds.

\textbf{Multiagent communication.} Modelling multiagent communication dates back to 1998 when Kinney et al.~\cite{kinney1998learning} proposed adaptive learning of multiagent communication strategy as a predefined rule-based control system. To scale up from rule-based systems, decentralized partially observable Markov decision process (DEC-PODMP) was used to model multiagent interaction with communication as a special type of action among agents~\cite{bernstein2002complexity,goldman2003optimizing}. Solving DEC-POMDP exactly is a NEXP-complete problem~\cite{bernstein2002complexity} requiring agents to remember the complete observation history for their policy. In recent works, a more compact representation of the history is often used for action selection. In~\cite{sukhbaatar2016learning,foerster2016learning}, agents maintain a memory variable and refer to it when act and speak. However, the centralized training process in these work needs channels with large bandwidth to pass gradients across different agents. Also, the messages in~\cite{sukhbaatar2016learning} are not discrete symbols, but continuous outputs of neural networks. Other extension of single agent value based algorithms to multiagent problems~\cite{lowe2017multi,foerster2016learning} usually suffers from non-stationariy induced by simultaneous updates of agents. 

While DEC-POMDP considers other agents as part of the environment and learns the policy as a mapping from local observation to action, LOLA in~\cite{foerster2018learning} learns the best response to evolving opponents. Yet, opponent/partners' real-time belief is not considered into policy. Interactive-POMDP (I-POMDP)~\cite{doshi2009monte,han2018learning} moves one more step forward by actually modeling opponents' mental states at the current moment and integrates other's belief into the agent's own policy. However, I-POMDP requires extensive sampling to approximate the nested integration over the belief space, action space and observation space, limiting its scalability. Because whenever the teacher sends a message, she thinks about not only the student's current belief but also current distractor sets, the value iteration process in I-POMDP needs to be repeated for every single game. In our algorithm, training only needs to be performed once for one pair of agents and reusable to all games. The Bayesian action decoder (BAD)-MDP proposed by Foerster et al.~\cite{foerster2018bayesian} also yields counterfactual reasoning in their belief update, but their method is more centralized in the testing process than ours. The BAD-agent is a super-agent controlling all other agents collectively. Deterministic partial policies can easily reveal agents' private information to the BAD-agent and make it public. Instead, our model doesn't depend on any implicit information flowing between agents during testing.
\section{Adaptive Emergence of Pragmatic Protocol}\label{sec:method}
\begin{figure}
    \centering
    \begin{subfigure}[b]{0.5\textwidth}
        \includegraphics[width=\textwidth]{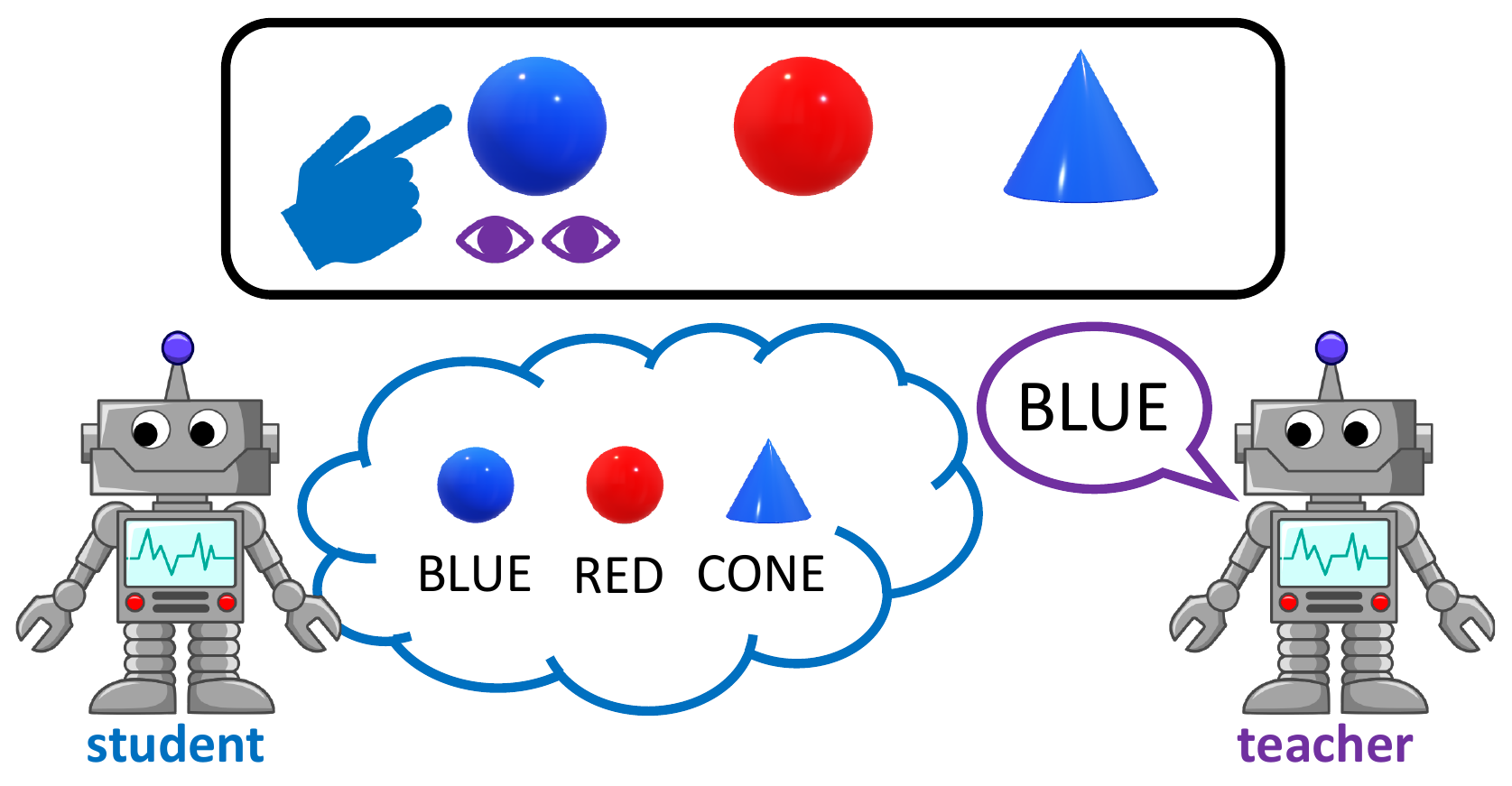}
        \caption{Referential Game}
        \label{fig:game}
    \end{subfigure}~
    \begin{subfigure}[b]{0.5\textwidth}
        \includegraphics[width=\textwidth]{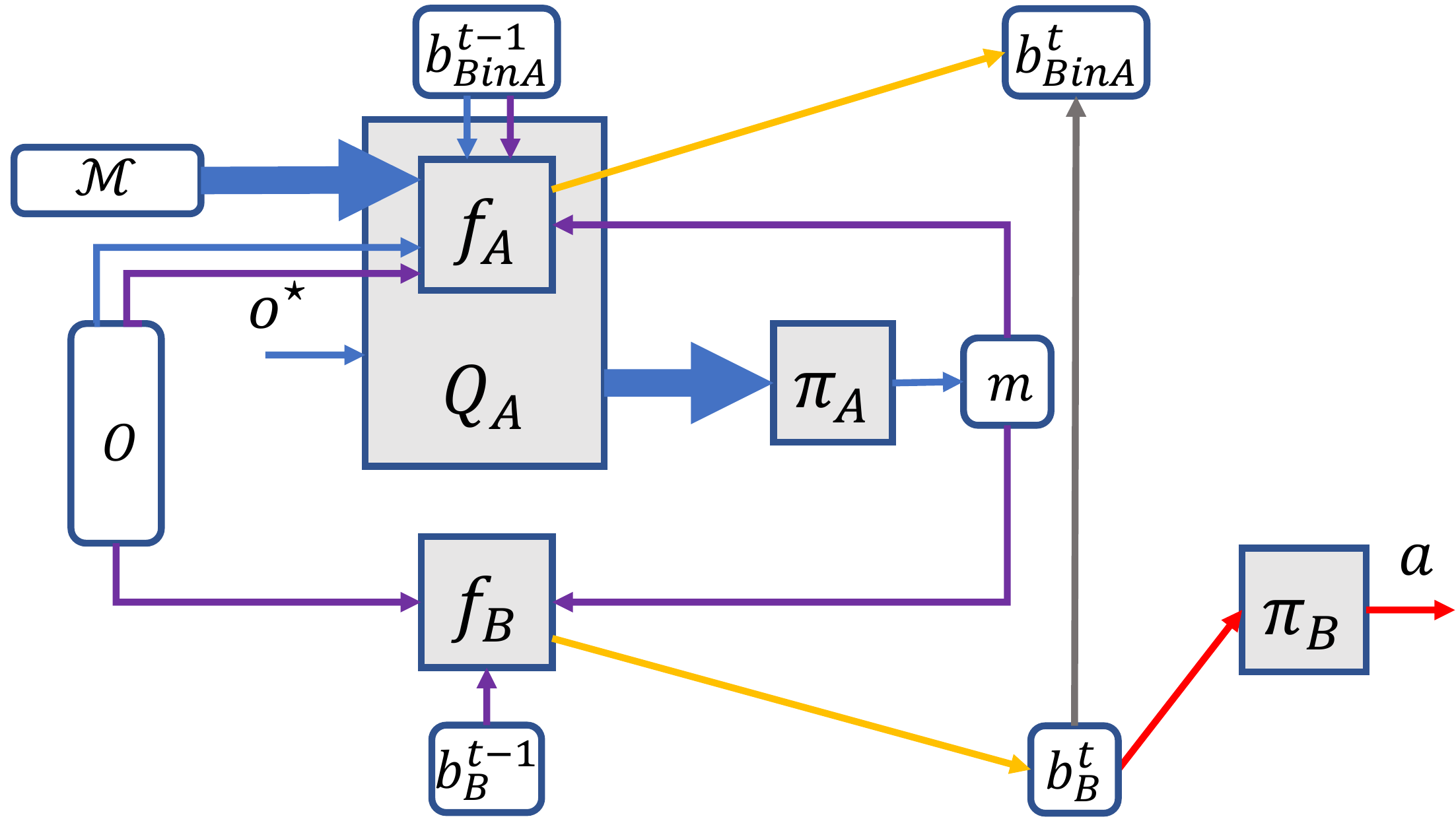}
        \caption{ToM Agents Interaction Pipeline}
        \label{fig:pipeline}
    \end{subfigure}
    \caption{(a) An example referential game. There are three objects, a blue sphere, a red sphere, and a blue cone. If a student hears “blue” from the teacher, he should be able to identify the blue sphere instead of the blue cone. (b) ToM Agents Interaction Pipeline. First, the teacher chooses a message according to the context and her prediction of the student's reaction (\textbf{\textcolor{ceruleanblue}{blue}} arrows). After a message is sent, the student updates his belief and the teacher updates her estimation of student's belief (\textbf{\textcolor{darkmagenta}{purple}} and \textbf{\textcolor{fluorescentorange}{orange}} arrows). Then, the student either waits or selects a candidate (\textbf{\textcolor{red}{red}} arrows). Only in the training phase, the actual student belief will be returned to the teacher (\textbf{\textcolor{charcoal}{gray}} arrow). Bold arrows stand for the whole message space being passed. Notice $f_A$ is part of the $Q_A$. $O$ and $b_{BinA}^{t-1}$ are passed to $f_A$ twice, for message selection and teacher's new belief estimation. Empty boxes are game and time variants; shadowed boxes are agents' mental structures. Notations are introduced in section \ref{sec:method} with $\theta$ omitted from subscripts.}\label{fig1}
    \vspace{-0.35cm}
\end{figure}

\begin{algorithm}[t!]
\footnotesize
\caption{Iterative Adaption Protocol Emergence}
\label{alg1}
\setlength{\columnsep}{-2.2cm}
\begin{multicols}{2}
\begin{algorithmic}[1]
   \small
    \STATE Randomly initialize $\theta_A, \theta'_A, \theta_B$\\
    \STATE No. candidates $K$\\
    \STATE Learning rate $\eta$, Batch size $N$
    \FOR{each phase}
        \FOR{$i \in \{A, B\}$}
            \STATE Initialize replay buffer $\mathcal{D} \leftarrow \emptyset$
            \WHILE{train agent $i$}
                \STATE $t = 1$
                \STATE Initialize $\mathcal{E} \leftarrow \emptyset$
                \REPEAT
                    \IF{$t = 1$} \label{algline:interaction_start}
                        \STATE Sample $O = \{\omega_1, ..., \omega_K\}$\\
                        \STATE Random select $o^{\star} = \omega_j$\\
                        \STATE Initialize $b_B^0, b_{BinA}^0$\\
                        as uniform distribution
                    \ENDIF
                    \STATE $m_t \sim \pi_{\theta_A}(\boldsymbol{\cdot}|O, o^{\star}, b_{BinA}^{t-1})$\\
                    \STATE $b_B^t = f_{\theta_B}(O, b_B^{t-1}, m_t)$\\
                    \STATE $a_t \sim \pi_{\theta_B}(b_B^t)$\\
                    \STATE $r_t = -c_{m_t} + \mathbf{1}(O[a_t] = o^{\star})$
                    \STATE $b_{BinA}^t = b_B^t$ \label{algline:interaction_end}
                    \IF{$i = A$}
                        \STATE $\mathcal{D} \leftarrow \mathcal{D} \cup \{(O, o^{\star}, b^{t-1}_{BinA}, m_t, b^t_B, r)\}$
                    \ELSE
                        \STATE $\mathcal{E} \leftarrow \mathcal{E} \cup \{(O,b^{t-1}_B, m_t, a_t, r)\}$
                    \ENDIF
                    \STATE $t \leftarrow t + 1$
                \UNTIL{$a_t \neq \Xi$}
                \IF{$i = A$} \label{algline:teacher_start}
                    \STATE Sample $\{(O, o^{\star}, b_{BinA}^{t-1}, m_t, b_{BinA}^t, r)\}_N \sim \mathcal{D}$\\
                    \STATE $\xi = r + \gamma\argmax_m Q_{\theta_A'}(O, o^{\star}, b_{BinA}^t, m)$\\
                    \STATE $L^Q = \frac{1}{N}\sum_N ||\xi - Q_{\theta_A}(O, o^{\star}, b_{BinA}^{t-1}, m_t)||^2$\\
                    \STATE $L^{Obv} = \frac{1}{N}\sum_N H(b_{BinA}^t, f_{\theta_A}\left(O, b_{BinA}^{t-1}, m_t)\right)$\\
                    \STATE $\theta_A \leftarrow \theta_A - \eta\nabla_{\theta_A}(L^Q + \lambda L^{Obv})$ \label{algline:teacher_loss}
                    \STATE Update $\theta_A' \leftarrow \theta_A$ periodically \label{algline:teacher_end}
                \ELSE \label{algline:student_start}
                    \STATE Compute $R_t = \sum_{k=t}^{|\mathcal{E}|} \gamma^{k-t}r_k$ for $t$ in ${1, ..., |\mathcal{E}|}$\\
                    \STATE $J = \frac{1}{|\mathcal{E}|}\sum_{t=1}^{|\mathcal{E}|} \log \pi_{\theta_B}\left(a_t|f_{\theta_B}(O, b_B^{t-1}, m_t)\right)R_t$\\
                    \STATE $\theta_B \leftarrow \theta_B + \eta\nabla_{\theta_B}J$
                \ENDIF \label{algline:student_end}
            \ENDWHILE
        \ENDFOR
    \ENDFOR
\end{algorithmic}
\end{multicols}

\end{algorithm}
\textbf{Emergence of Pragmatic Protocol:}
Our goal is to learn a protocol for agent $A$ and $B$ so that they can communicate with the contextual information being considered. To avoid tracking the message history, which scales exponentially with the time, we use beliefs as sufficient statistics for the past. Hence, ToM can be embodied as estimating partner's current and future belief, then choose the most ideal action to manipulate them as needed. In the referential game, since the teacher knows the target, only the student holds a belief, $b_B$, about the target. Utilizing the obverter technique~\cite{bogin2018emergence,choi2018compositional}, we let the teacher holds a belief $b_{BinA}$ as her estimation of student's belief. As $b_{BinA}$ is an estimation of the student's belief, it should be a belief over belief, i.e. a distribution over a distribution over the candidates. Since a distribution is a continuous random variable, distribution over a continuous variable can be represented as a set of particles. However, to avoid the complexity we only use one particle to approximate. That is, $b_{BinA}$ is still a distribution over the candidates. This is reasonable because the belief update process is deterministic for rational agents following the Bayesian rule~\cite{vogel2013emergence,fisac2017pragmatic}. Given $b^0_B$ a uniform distribution over candidates, $P(b^t_B)$ is a single mode distribution and can be approximated with a particle.

Before speaking, teacher traverses all messages and predicts the student's new belief after receiving each message. She then sends the message leading to the most optimal student's new belief. Hearing the message, student updates his belief and takes action. This process is visualized in figure \ref{fig:pipeline} and formalized in algorithm \ref{alg1} line \ref{algline:interaction_start} to \ref{algline:interaction_end}. The recursive mutual modeling in ToM is integrated within the belief update process. $f_{\theta_i}, i \in \{A, B\}$ are belief update functions parameterized by $\theta_i$, taking in candidates, current belief, message and returning a new belief. The beliefs in our model are semantically meaningful hidden variables in teacher's Q-function and student's policy network, as the student directly samples an action according to his belief. The evolving of the belief update function reflects the protocol dynamics between the agents.
Within $f$, we code in the Bayesian rule with the likelihood function varying across different training phases. Our implementation detail can be found in section \ref{sup:structure}. In each phase, we first train the teacher for a fixed student, then adapt the student to the teacher.

\textbf{Difference from multiagent Q-learning:} Our algorithm considers both the physical state and agent's mental state in the value function, and has a dynamic belief update function. Moreover, since agents are never trained simultaneously, our algorithm doesn't suffer from non-stationarity~\cite{lowe2017multi,foerster2016learning}.

\textbf{Teacher:}
The teacher selects messages according to her Q-values and belief update function. We use $f_{\theta_A}(O, b, m)$ to denote teacher's belief update function, which takes in the candidates set, current belief estimation and a message. The return value of this function is a new belief estimation $b' \in \Delta O$. $\Delta O$ represents all probabilistic distributions over the candidates. This function can be parameterized as a neural network with weighted candidates encoding and messages as inputs and softmax as the output layer. The return value of the belief update function is directly fed into the Q-function. In practice, we implement it as a submodule of the Q-net. That is, the output of the belief update function is used in $A$'s Q-function and to predict student's belief in next step during testing. The teacher chooses messages according to her Q-value following equation \ref{eq:softmax}.
\begin{align}
    \pi_{\theta_A}(m|O, o^{\star}, b) = \dfrac{\exp\left(\beta Q_{\theta_A}(O, o^{\star}, b, m)\right)}{\sum_{m' \in \mathcal{M}}\exp\left(\beta Q_{\theta_A}(O, o^{\star}, b, m')\right)} \label{eq:softmax}
\end{align}
\begin{align}\label{eq:teacher_q}
    Q&_{\theta_A}(O, o^{\star}, b, m) = \mathbf{E}_{a \sim\pi_{\theta_B}\big(f_{\theta_B}(O, b, m)\big)}\Big[\mathbf{1}(O[a] = o^*) + \\ \nonumber
    &\mathbf{1}(a = \Xi)\max_{m'} Q_{\theta_A}\big(O, o^{\star}, f_{\theta_B}(O, b, m), m')\big)\Big] - c_m
\end{align}
Equation \ref{eq:teacher_q} defines the teacher's Q-function. $\mathbf{1}(O[a]=o^*)$ indicates whether the student makes a correct prediction. $\mathbf{1}(a = \Xi)$ indicates if the game is still ongoing. Student's belief $b$ is the state of teacher's MDP. Since student's actions determine the game states, the expectation is over the student's policy. By definition, the teacher's Q-function relies on student's policy and belief update function. She has no access to these student's functions, but since we never train the teacher and student simultaneously, the expectation can be approximated through Monte-Carlo (MC) sampling. To form a protocol, agent $A$ needs to learn two functions, her belief update function $f_{\theta_A}$ and $Q_{\theta_A}$. In the training phase, every time the student receives a message, he returns his new belief $b_B^t$ to the teacher. During testing, she needs to use the output of $f_{\theta_A}$ to approximate student's new belief. We train $f_{\theta_A}$ by minimizing the cross-entropy, $H$, between $b_B^t$ and teacher's prediction, denoted as $L^{Obv}$, the obverter loss. Teacher's Q-function is learned with Q-learning \cite{watkins1989learning}. The $\lambda$ in line \ref{algline:teacher_loss} controls the scale of the two losses.

\textbf{Student:}
We directly learn the belief update function and policy of the student through the REINFORCE algorithm \cite{williams1992simple}. In the referential game, student's policy is quite simple. If his belief is certain enough, he will choose the target based on his belief; otherwise, wait for further messages. The output of the policy network is a distribution with $|O| + 1$ dimensions. The last dimension is a function of the entropy of the original belief. If the belief is uncertain, this value will be dominant after normalization. $f_{\theta_B}$ has the same structure as $f_{\theta_A}$. $f_{\theta_B}$ and $\pi_{\theta_B}$ can be parameterized as an end-to-end trainable neural network, with the candidates encoding, original belief and received a message as the input and returning an action distribution. 

\textbf{Adaptive Training:} The whole training process can then be
summarized as Algorithm \ref{alg1}. Both the teacher and student are adaptively trained to maximize their expected gain defined in Eq.~(\ref{eq:gain}). The training details for the teacher and the student are illustrated in Line \ref{algline:teacher_start}-\ref{algline:teacher_end} and Line \ref{algline:student_start}-\ref{algline:student_end} of Algorithm \ref{alg1} respectively.
\section{Experiments} \label{sec:exp}
\begin{figure*}[t]
    \centering
    \begin{subfigure}[b]{0.3\textwidth}
        \includegraphics[width=\textwidth]{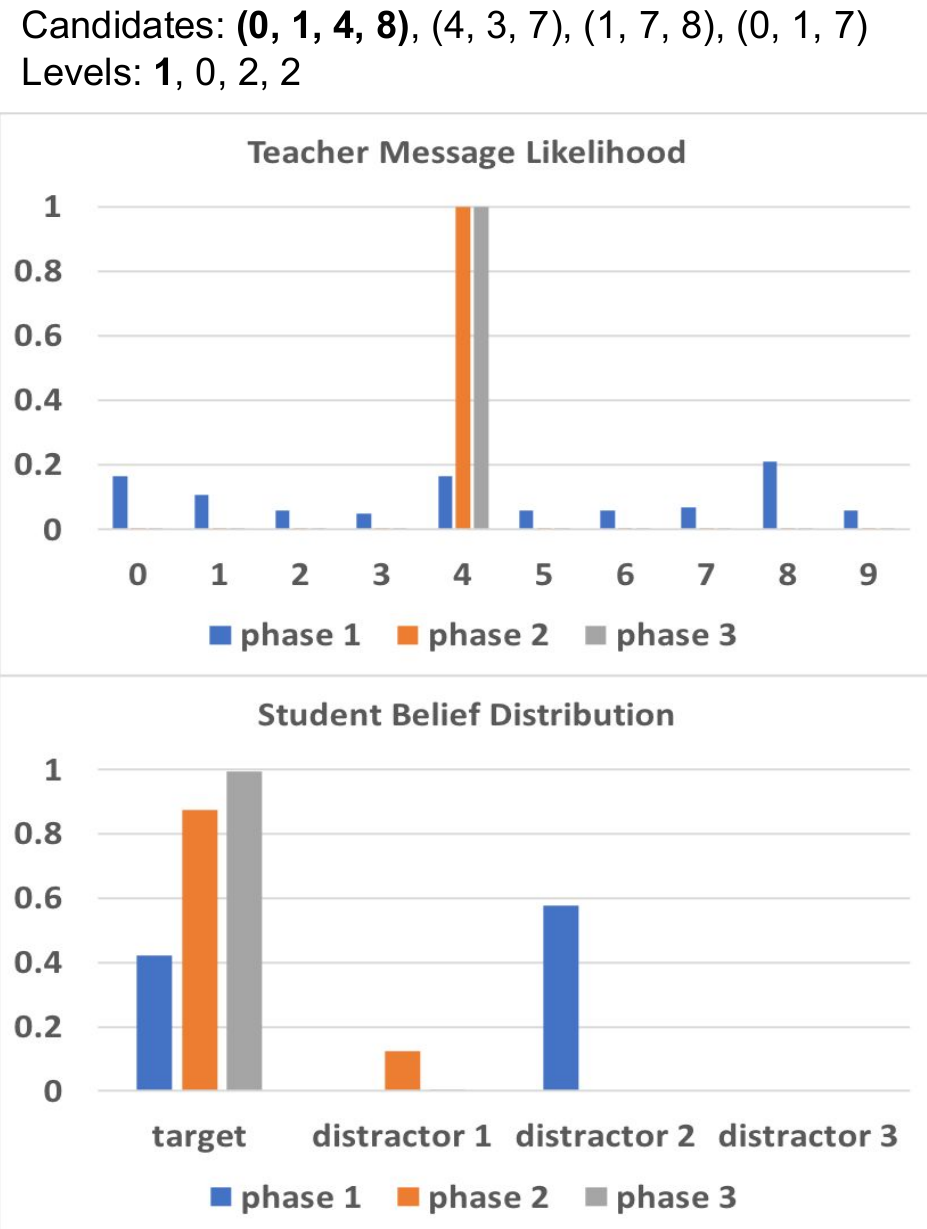}
    \end{subfigure}~
    \begin{subfigure}[b]{0.6\textwidth}
        \includegraphics[width=\textwidth]{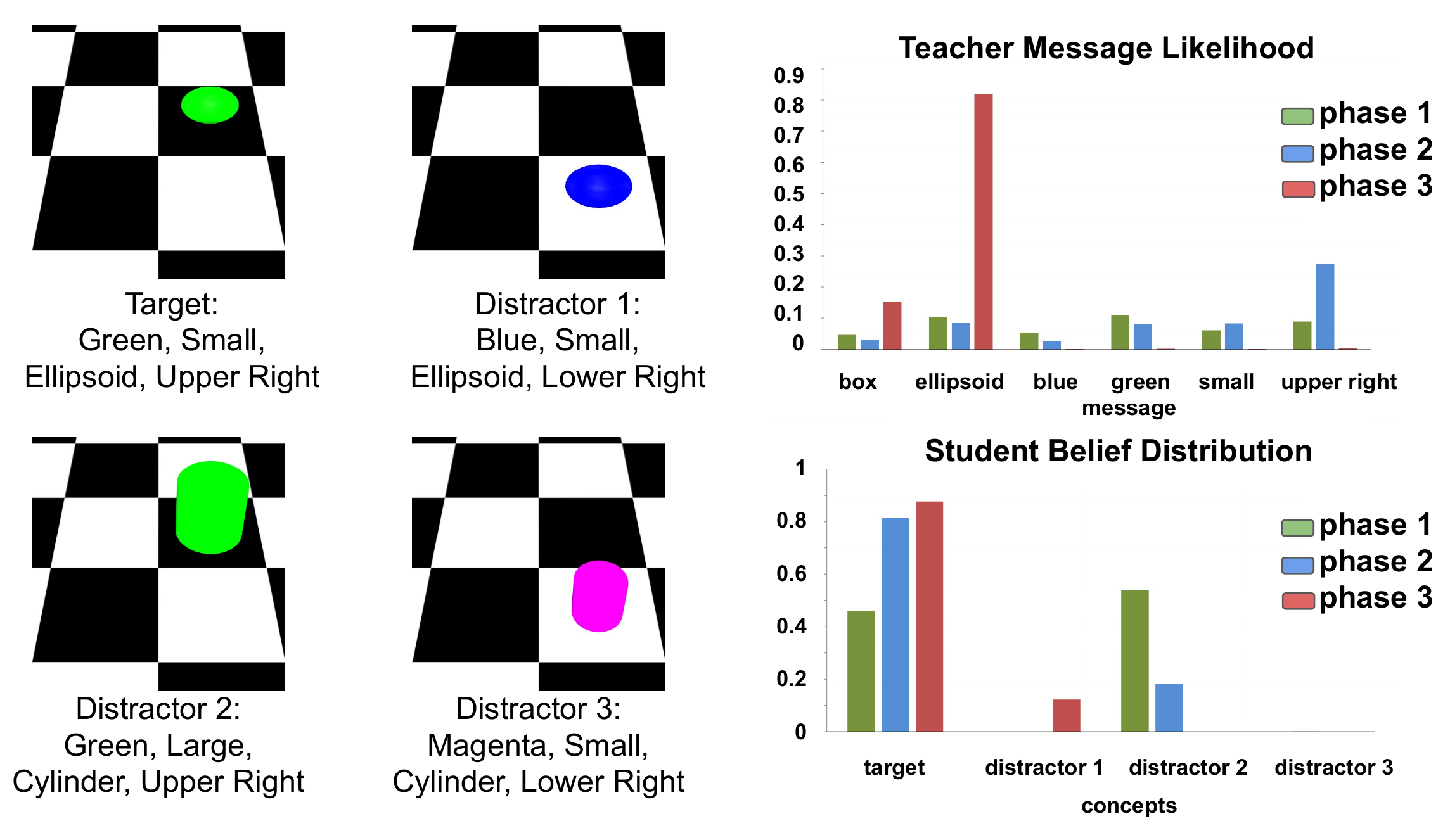}
    \end{subfigure}
    \caption{\small{4 distractors referential game example. Number set on the left (candidates listed in the title with the target in bold fonts) and 3D objects on the right. Due to the space limit, we only show the message distribution for the target and student's new belief after receiving the most probable message. As for the teacher's message distribution for distractors, all probability weights concentrate on the unique identifiers after the first phase of training. Student's belief illustrates that teacher's most probable message, though consistent with multiple candidates, can successfully indicate the target with more confidence as training goes. In general, both agents' behavior becomes more certain, and the certainty coordinates.}}
    \label{fig:RGexps}
    \vspace{-10pt}
\end{figure*}
\begin{table}
    \centering
    \setlength{\tabcolsep}{3pt}
    \caption{The \textbf{first three rows} are the total accuracy, while the \textbf{second three rows} measure the accuracy for difficult games (numbers in percentage). We define game difficulty with average cosine similarity between the target and the distractors. The larger the cosine similarity, the harder the game is. We report the accuracy for top $10\%$ hard games. L.A. stands for Lazaridou et el. To verify the generalization of our algorithm, we form training sets with $70\%$ of all instances and test using the rest $30\%$ unseen instances. The performance doesn't exhibit noticeable degradation: \textbf{99.1 $\pm$ 0.1$\%$} and \textbf{93.0 $\pm$ 1.7$\%$} for 4 and 7 candidates number set respectively (comparable with the No. Set results in the third line). Mean and std calculated using 3 different random splits, 2 experiments per split.}
    \begin{tabular}{cc|c|cc}
        \toprule
        No. Set& 3D objects & $\leftarrow 4$ $|$  7 $\rightarrow$& No. Set & 3D Objects \\
        \midrule
        79.1 $\pm$ 3.3 & 86.9 $\pm$ 4.1 & \small{L. A. [2018]} & 64.2 $\pm$ 6.0 & 77.3 $\pm$ 2.8\\
        96.8 $\pm$ 0.2 & 97.0 $\pm$ 0.5 & \small{L. A. [2017]} & 80.8 $\pm$ 3.1 & 88.2 $\pm$ 1.7 \\
        \textbf{98.9 $\pm$ 0.1} & \textbf{99.6 $\pm$ 0.2} & Pragmatics & \textbf{93.2 $\pm$ 1.0} & \textbf{97.4 $\pm$ 1.2}\\
        \midrule
        79.3 $\pm$ 3.1 & 86.9 $\pm$ 4.5 & \small{L. A. [2018]} & 67.2 $\pm$ 5.8 & 77.2 $\pm$ 2.6\\
        91.5 $\pm$ 0.4 & 88.0 $\pm$ 1.9 & \small{L. A. [2017]} & 66.2 $\pm$ 3.1 & 68.2 $\pm$ 2.7\\
        \textbf{98.1 $\pm$ 0.3} & \textbf{98.8 $\pm$ 0.3} & Pragmatics & \textbf{88.3 $\pm$ 0.6} & \textbf{94.1 $\pm$ 2.3}\\
        \bottomrule
    \end{tabular}
    \label{tab:refG_results}
    \vspace{-15pt}
\end{table}
We evaluated our algorithm with two datasets, number set and 3D objects, and played referential games with four or seven candidates. The number set is a symbolic dataset, with an instance as a set of categorical numbers. For example, $[(1, 2, 3, 9), (1, 2, 4), (2, 3), (3, 4, 5)]$ consists a referential game with four candidates. Notice that the numbers are merely symbols without numerical order. If there are four candidates, we randomly choose numbers from 0 to 9, with maximum four numbers in a set; if seven candidates, we choose from 0 to 11, with maximum five numbers in a set.  Each set is encoded by multi-hot encoding. There are 385 and 1585 different possible number sets, consisting up to $9.0\times 10^9$ and $4.9\times 10^{18}$ different games with four and seven candidates. Number sets make a generic referential game prototype, where each instance can be disentangled into independent attributes perfectly. To verify the generality of our algorithm on more complicated candidates, we used MoJoCo physical engine to synthesize RGB images of resolution $128 \times 128$ depicting single 3D object scenes. For each object, we pick one of six colors (blue, red, yellow,
green, cyan, magenta), six shapes (box, sphere, cylinder, pyramid, cone, ellipsoid), two sizes and four locations, resulting in 288
combinations. In every game, candidates are uniformly sampled from the instances space. We use a message space with the same size as the number of attributes appeared in the dataset, i.e., 10 or 12 for number set, and 18 for 3D objects. In every game, we only allow one round of communication with one message. To prevent collusion using trivial position indicator, candidates are presented to the agents in different orders.

\subsection{Referential Game with Symbolic and Pixel Input}
For number set, we encode candidates with multi-hot vectors and messages with one-hot vectors. For 3D objects, we used a convolutional neural network (CNN) to extract features of the candidates. For all datasets, we manually generated 600k and 100k mutually exclusive games for training and testing. Namely, the same instances can appear in both datasets, but not any identical candidates combinations. To win a game, the same instance needed to be handled differently given different contexts, so, as long as the games are exclusive between training and testing sets, sharing instances won't cause over-fitting. To test the robustness, we also report results using exclusive testing instances in table \ref{tab:refG_results}. We compared the pragmatic protocol developed using our algorithm against previous works on referential game \cite{lazaridou2016multi,lazaridou2018emergence}. Both utilized RL to train a protocol, but neither modeled recursive mind reasoning between agents. In \cite{lazaridou2016multi}, only the teacher considers the context, while no context is included in \cite{lazaridou2018emergence}. We trained our model for 3 phases, with 20k iterations for each phase and switch the training agent in the middle of every phase. Both benchmarks were trained for 100k iterations. Since there is no official code released by the authors, we implemented their model by ourselves and did thorough hyper-parameter grid search. Results shown in table \ref{tab:refG_results}. Our experimental results in all settings are significantly better than both of the benchmarks. Using the paired T-test, the one-tail $p$-value is smaller than 0.001 for all settings in table\ref{tab:refG_results}. We found that even with simpler representation, number set games are more difficult than 3D objects, because we don't have any limitations generating the instances in number sets. 3D objects, on the other hand, form special cases of number sets, as some attributes can never coexist. E.g. a shape cannot be a sphere and a cone simultaneously.

\subsection{Connection with RTD} \label{sec:exp_rtd}
\begin{SCfigure}
    \centering
    \footnotesize
    \includegraphics[width=0.5\textwidth]{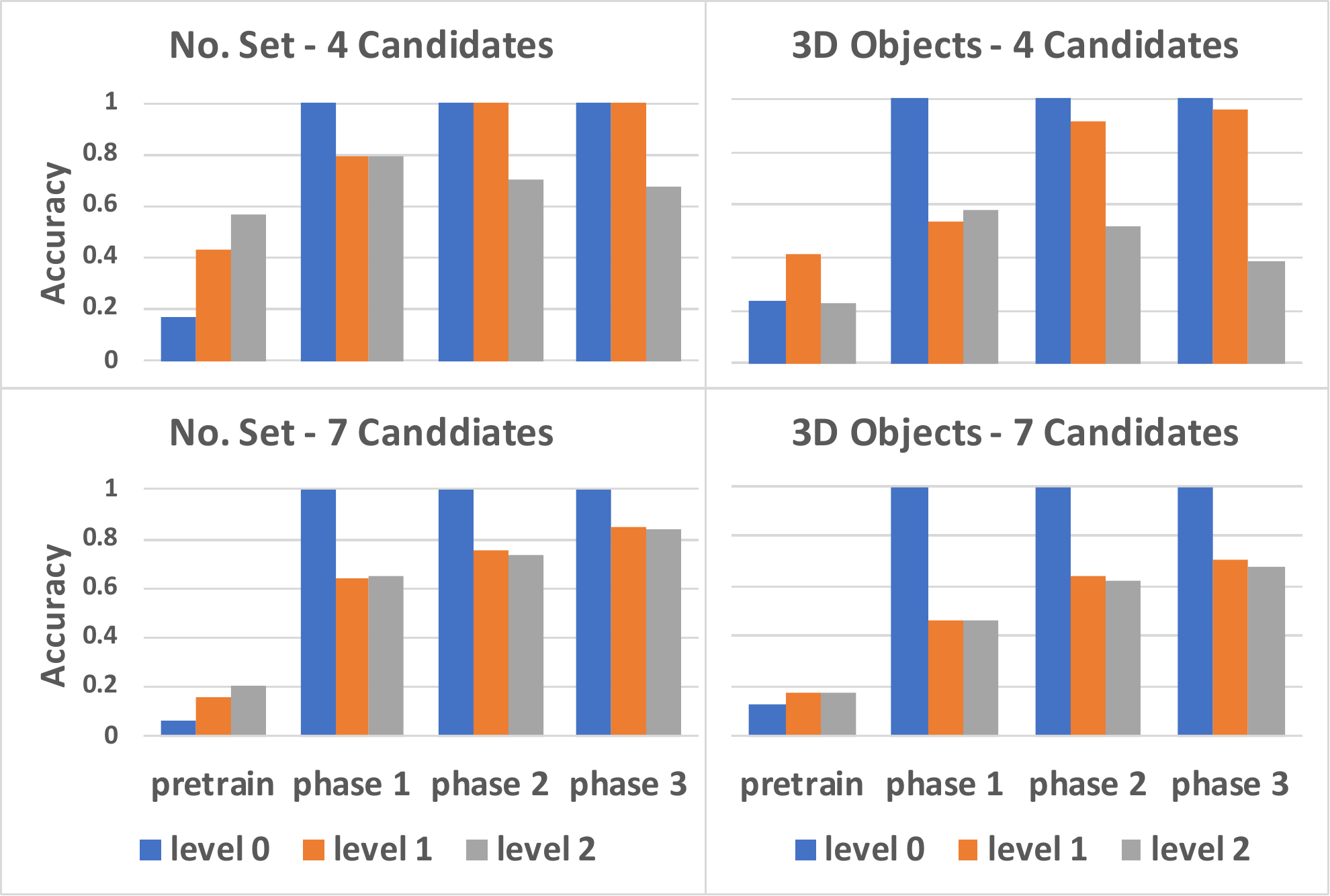}
    \caption{Finishing one phase of adaptation, agents can correctly identify level 0 targets. Namely, the teacher always sends a unique identifier as long as there is one. After phase 2, the student knows how the teacher will teach level 0 targets. Thus, he can prune out all level 0 candidates if he doesn't receive their unique identifiers, leaving level 1 candidates the “simplest”. The decrease of level 2 accuracy among 4 candidates might be caused by the lack of level 3 targets. Only $3\%$ of the number set and $0.02\%$ 3D Objects games have level 2 targets, so agents have no motivation to explore for a higher level of reasoning.}
    \label{fig:RTDaccuracy}
\end{SCfigure}

The iterative adaptive idea of our algorithm is similar to the definition of RTD, which measures the number of examples needed for concept learning between a pair of cooperative and rational agents~\cite{chen2016recursive}. We included the formal definition of RTD in section \ref{sup:RTD} of the appendix. Intuitively, in a concept class, there is a subset of concepts which are the simplest to learn i.e. has the minimum sized teaching set among all concepts. One can first learn those concepts and remove them from the concept class. Now, for the remaining set of concepts, one can recursively learn the simplest concepts and so on. The teaching complexity of this learning schema lower bounds classic teaching dimension~\cite{doliwa2014recursive}. In every phase of our iterative training, the agent learns to identify the optimal teaching set for the “simplest” remaining candidates. In our experiments, candidates identifiable with a unique message are the simplest. If a candidate becomes the simplest after $k$ times of removal, then we call it a level $k$ candidate. To better illustrate the connection, we show two example referential games in figure \ref{fig:RGexps} and the accuracy improvement after each phase of training in figure \ref{fig:RTDaccuracy}. We can see from figure \ref{fig:RTDaccuracy} that after one phase of training all level 0 targets can be perfectly identified. Thus, the student, if shown the four 3D objects in figure \ref{fig:RGexps}, will know that the teacher will send “Blue” for distractor 1, “Large” for distractor 2 and “magenta” for distractor 3. Hence, “Upper Right” and “Ellipsoid”, though consistent with multiple objects, must indicate the target. The accuracy for higher-level targets in figure \ref{fig:RTDaccuracy} keeps increasing as they become uniquely identifiable after lower-level targets are pruned out. We can observe the emergence of pragmatics from these results. From the student's perspective, the messages from the teacher are no longer merely comprehended by their literal meanings, and from the teacher's perspective, she selects the most helpful message to teach. In the 4-candidate scenario, most questions with level 0 and 1 are correctly answered, a similar capability shown in human one-shot referential game study~\cite{bergen2012s}.

Notice that RTD is derived under the assumption that both agents only decipher the candidates as sets of discrete attributes. To eliminate the usage of other hidden patterns in the candidates, we need to pretrain the agents to ground messages to instances attributes. This can be easily achieved by initializing agents' belief update function as Bayesian belief update. That is, before running algorithm \ref{alg1}, we train the belief neural network $f$ with cross-entropy loss between the generated new belief and ground truth Bayesian belief. Afterward, every message grounds to an attribute.

Bayesian pretraining provides human decipherable examples for failure cases. Most of our model's mistakes are on targets with $RTD > 1$ or $RTD = 1$ but requiring high ($\geq 2$) level unique identifiers (hard games). Since we only allow one message, targets with $RTD > 1$ are theoretically impossible to be certainly identified, even between cooperative agents with ToM. As for the hard games, their relatively low frequency in the training set may impinge the acquiring of high-level best response. Failure to handle a certain type of scenarios is a common empirical defect of the current RL algorithms, in our case, the hard games. Another benefit of Bayesian pretrain is that the initial message grounding is decipherable to human. In the next section, we show this interpretability can be preserved by our algorithm.

\subsection{Stability of the Protocol}\label{sec:stable}
We also explored how much the communication changes after the mutual adaptation process. Suppose the agents are initialized with human-understandable message groundings, ideally, we want the emerged protocol preserves its human interpretability. To test this property, we give the teacher and student different but equivalent candidates. Namely, we randomly generate a one-to-one mapping from attributes to attributes, such as replacing all 1 with 7 in number sets or replacing all red with blue in 3D objects. After the teacher sends a message to the student, both the message and candidates are converted with the same mapping before presented to the student. The converted candidates form an equivalent game for the original one. For example, if the teacher sees $[(0, 1, 5, 8), (0, 4, 6), (4, 8, 9), (0, 1, 4, 5)]$ and sends 4. Converted by a mapping which adds 1 to all attributes, the student gets $[(1, 2, 6, 9), $ $(1, 5, 7), (5, 9, 0), (1, 2, 5, 6)]$ and 5.

\begin{SCtable}
    \centering
    \caption{All models are pretrained with Bayesian belief update. The three rows above are tested with agents seeing different but equivalent candidates, while the three rows below is tested with agents sees identical candidates. The larger the performance difference is, the more the protocol diverges from the initial grounding.}
    \setlength{\tabcolsep}{3pt}
    \begin{tabular}{cc|c|cc}
        \toprule
        No. Set& 3D Objects & $\leftarrow 4$ $|$ 7 $\rightarrow$& No. Set & 3D objects \\
        \midrule
        59.5 $\pm$ 0.4 & 53.3 $\pm$ 1.8 & L. A. [2018] & 35.0 $\pm$ 0.2 & 37.7 $\pm$ 1.8\\
        52.3 $\pm$ 2.9 & 60.8 $\pm$ 2.9 & L. A. [2017] & 33.3 $\pm$ 2.0 & 46.0 $\pm$ 1.4\\
        \textbf{97.5 $\pm$ 0.4} & \textbf{98.7 $\pm$ 0.5} & Pragmatics & \textbf{73.9 $\pm$ 0.3} & \textbf{84.2 $\pm$ 0.1}\\
        \midrule
        81.5 $\pm$ 2.6 & 83.7 $\pm$ 3.4 & L. A. [2018] & 74.9 $\pm$ 0.5 & 75.3 $\pm$ 2.9\\
        \textbf{98.6 $\pm$ 1.0} & 98.1 $\pm$ 0.2 & L. A. [2017] & \textbf{93.0 $\pm$ 0.5} & 91.2 $\pm$ 0.5\\
        98.1 $\pm$ 0.3 & \textbf{99.7 $\pm$ 0.3} & Pragmatics & 88.2 $\pm$ 0.8 & \textbf{91.2 $\pm$ 0.1}\\
        \bottomrule
    \end{tabular}
    \label{tab:stable_results}
    \vspace{-10pt}
\end{SCtable}

We pretrain agents with Bayesian belief update separately, then train them together without equivalent games but test them with equivalent games. See table \ref{tab:stable_results} for the results. We can see that our algorithm preserves human interpretability the most. The iterative adaptation contributes to stability because, in every phase, one agent is fixed while optimizing the other. Hence, the evolving of the protocol is not arbitrary and will maintain the effective part of the existing protocol while improving the rest. This property can facilitate human-robot communication, as we only need to provide natural language grounding to robots, and they can self-evolve to take the best advantage of this grounding without developing human undecipherable protocols.

\subsection{Global Mapping and Local Selection}
\begin{SCtable}
    \centering
    \setlength{\tabcolsep}{3pt}\caption{A message is valid if it corresponds to one of the attributes of a novel target. Std calculated with 3 different random splits.}
    \begin{tabular}{c|cc|cc}
        \toprule
       &\multicolumn{2}{c|}{Valid \%}& \multicolumn{2}{c}{Accuracy}\\
        &4 & 7 & 4 & 7\\
        \midrule
        Pragmatics & \textbf{98.8 $\pm$ 0.29} & \textbf{98.1 $\pm$ 0.36} & \textbf{98.6 $\pm$ 0.08} & 87.9 $\pm$ 0.61\\
        L.A. [2017] & 74.5 $\pm$ 5.69 & 79.1 $\pm$ 4.73 & 97.0 $\pm$ 0.02 & \textbf{88.8 $\pm$ 0.25}\\
        \bottomrule
    \end{tabular}
    \label{tab:novel}
\end{SCtable}

\begin{figure}[ht]
    \centering
    \begin{subfigure}[b]{0.4\textwidth}
    \includegraphics[width=\textwidth]{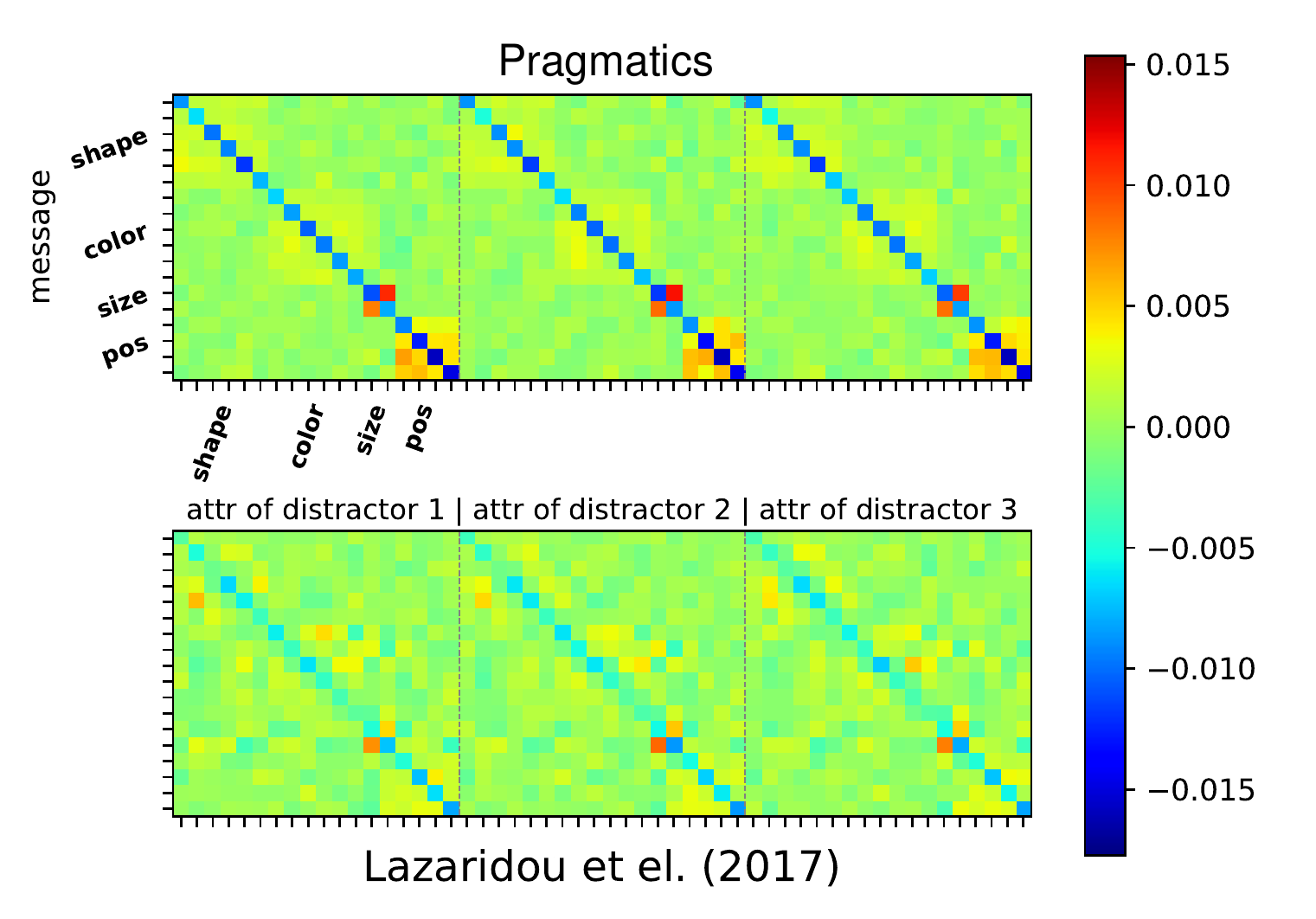}
    \caption{4 distractors}
    \end{subfigure}
    \begin{subfigure}[b]{0.59\textwidth}
        \includegraphics[width=\textwidth]{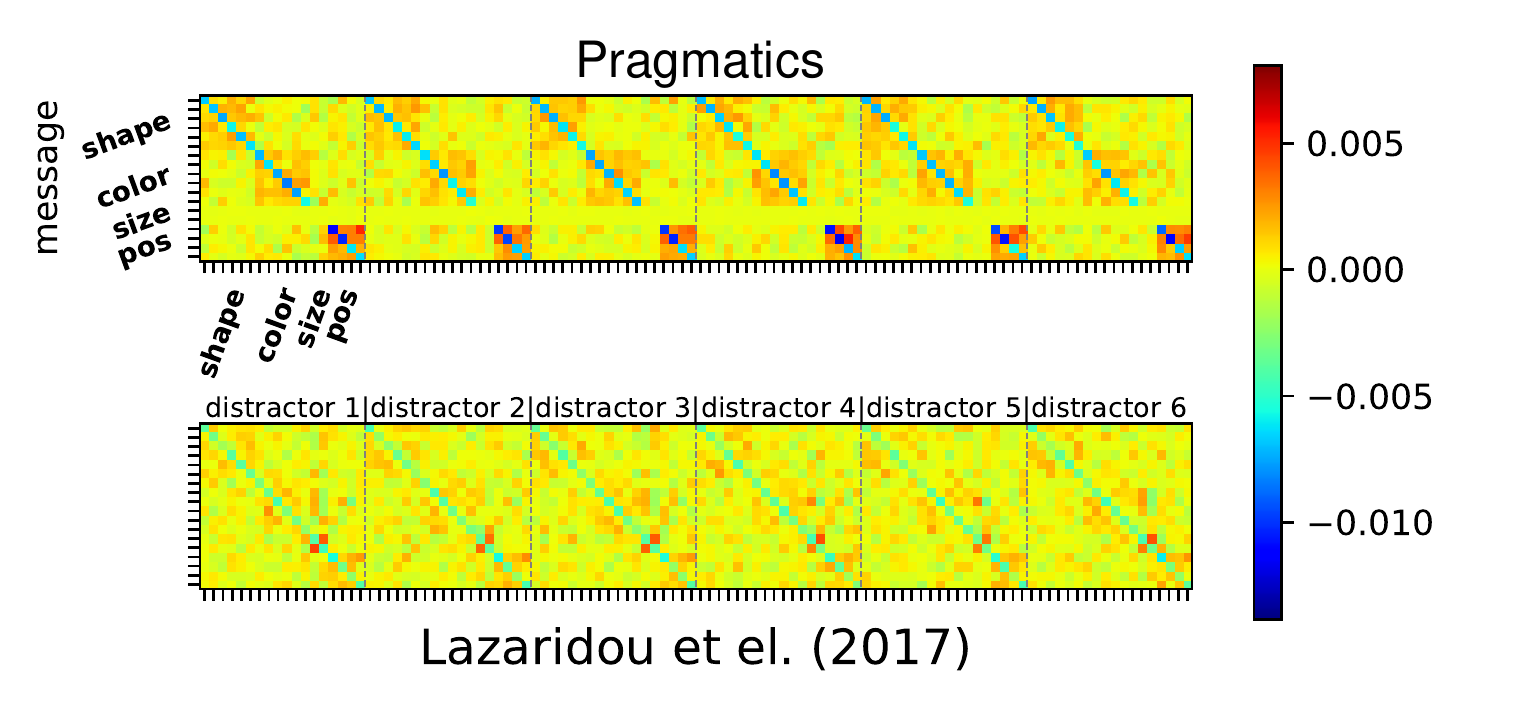}
    \caption{7 distractors}
    \end{subfigure}
    \caption{Covariance between messages and distractor attributes given the same target but different distractors. In both experiments, our algorithm has a more clear pattern. The blue diagonals show that if an attribute appears among distractors the teacher tends to avoid the corresponding message. The brighter sub-squares illustrate that messages are significantly influenced by distractors' attributes in the same category. Namely, color messages have stronger covariance with color attributes than with shapes, sizes or positions. Notice that size messages are seldom used in 7-distractor game, thus their covariance with distractors is close to 0. Better be viewed in high resolution, e.g. using Adobe Acrobat Reader.}
    \label{fig:cov}
    \vspace{-15pt}
\end{figure}
The core of pragmatics is the consideration of the context while comprehending the language. In this experiment, we want to show that the teacher using a pragmatic protocol can learn global mappings from instances to messages and select messages dynamically according to the context. First, we test global mapping by checking if messages used by the teacher for a target are consistent with this target's attributes. Then we evaluate message selection through seeing whether the same target yields different messages given different distractors. To make the message grounding easier to understand, we still pretrain agents with Bayesian belief update. We took 80\% of the instances to generate training data. In the testing phase, the rest 20\% of images are used as targets with 3/6 distractors randomly selected from all images. Results in table \ref{tab:novel} justify that the pragmatic protocols achieve the best balance between message validity and referential accuracy. Then we calculated covariance between messages and distractors attributes given the same target but changing distractors. For a target, the covariance is calculated using 100 games. In figure \ref{fig:cov}, we visualized mean covariance for 58 targets. Since the teacher doesn't have access to distractors in \cite{lazaridou2018emergence}, we only compared our model and \cite{lazaridou2016multi}.
\section{Conclusion}
In this paper we propose an end-to-end trainable adaptive training algorithm integrating ToM in multi-agent communication protocol emergence. The pragmatic protocol developed by our algorithm yields significant performance gain over non-pragmatic protocols. With human knowledge pre-grounding, the teaching complexity using the pragmatic protocol approximates RTD. Our algorithm incorporates a global mapping from instances to messages with a local message selection mechanism sensitive to the context. In future research, we hope to generalize the referential game to a new communicative learning framework, where students, instead of learning from data or an oracle, learn from a helpful teacher with ToM. We also plan to apply our algorithm to more generic communication settings, where agents have more symmetric roles. Namely, we have agents each holding some information unknown to the group and need communication to accomplish a task. We also want to relax the need of exchanging beliefs directly in the training phase, replacing it with discrete feedback requiring smaller channel bandwidth.

\bibliographystyle{IEEEtran}
\bibliography{reference}
\newpage
\begin{appendices}
\section{Recursive Teaching Dimension}\label{sup:RTD}
In computational learning theory, the teaching dimension (TD) measures the minimum number of examples required to uniquely identify all concepts within a concept class. We draw a comparison between identifying a concept using examples and identifying the target using messages. Formally, given an instance space $X = \{1,...,n\}$ and a concept class $\mathcal{C} \subseteq \{0,1\}^n$, a teaching set for a concept $c \in \mathcal{C}$ with respect to $\mathcal{C}$ is a subset $S \subseteq X$ such that $\forall c' \neq c \in \mathcal{C} \quad \exists x \in S$ s.t. $c(x) \neq c'(x)$. Intuitively, the teaching set is a subset of instance space that can uniquely identify $c$ in $\mathcal{C}$. TD of $c$ in $\mathcal{C}$, denoted by $TD(c, \mathcal{C})$, is the size of the smallest teaching sets. TD of the whole concept class is $TD(\mathcal{C}) = \max_{c\in \mathcal{C}}TD(c, \mathcal{C})$. However, this definition of TD often overestimates the number of examples needed for cooperative agents, who teach and learn using "helpful" examples. For instance, the game in figure \ref{fig:game} in the main text has TD = 2, but all concepts can be taught with one message.

The recursive teaching dimension (RTD), a variation of the classic TD, can model the behavior between cooperative agents. Define the teaching hierarchy for $\mathcal{C}$ as the sequence $((\mathcal{C}_1,d_1), ..., (\mathcal{C}_h,d_h))$ such that for all $j \in \{1,...,h\}$, $\mathcal{C}_j = \{ c \in \overline{\mathcal{C}}_j| d_j = TD(c, \overline{\mathcal{C}}_j) \leq TD(c', \overline{\mathcal{C}}_j), \forall c' \in \overline{\mathcal{C}}_j \}$ where $\overline{\mathcal{C}}_1=\mathcal{C}$ and for all $i \in \{1,...,h-1\}$, $\overline{\mathcal{C}}_{i+1} = \mathcal{C}\backslash(\mathcal{C}_1\bigcup ... \bigcup \mathcal{C}_i)$. RTD of $c$ in $\mathcal{C}$ is defined as $RTD(c, \mathcal{C}) = d_j$. RTD of $\mathcal{C}$ is $RTD(\mathcal{C}) = \max\{d_j|1 \leq j \leq h\}$. Intuitively, one can first learn simplest concepts, $\mathcal{C}_i$, and remove them from the concept class. Then, for the remaining set of concepts, $\overline{\mathcal{C}}_{i+1}$, recursively learn the simplest concepts and so on. RTD lower bounds the classic teaching dimension, i.e. $RTD(\mathcal{C}) \leq TD(\mathcal{C})$ \cite{doliwa2014recursive}. Also, $RTD(\mathcal{C})$ measures the worst-case number of labeled examples needed to learn any target concept in $\mathcal{C}$, and the teaching hierarchy can be derived by the teacher and student separately without any communication \cite{chen2016recursive}.

\section{Network Structure} \label{sup:structure}
Our network consists of 3 modules: a belief update network, teacher's Q-net and student's policy network. The teacher and student share the same structure for the belief update network but initialized differently. The output of the belief update network will feed into the Q-net or policy network for message and action selection. Figure \ref{fig:pipeline} shows the structure of our models. When dealing with image inputs, we have an additional perceptual module to process the visual inputs. We use a convolutional neural network to extract features from images. 

\begin{figure}
    \centering
    \includegraphics[scale = 0.3]{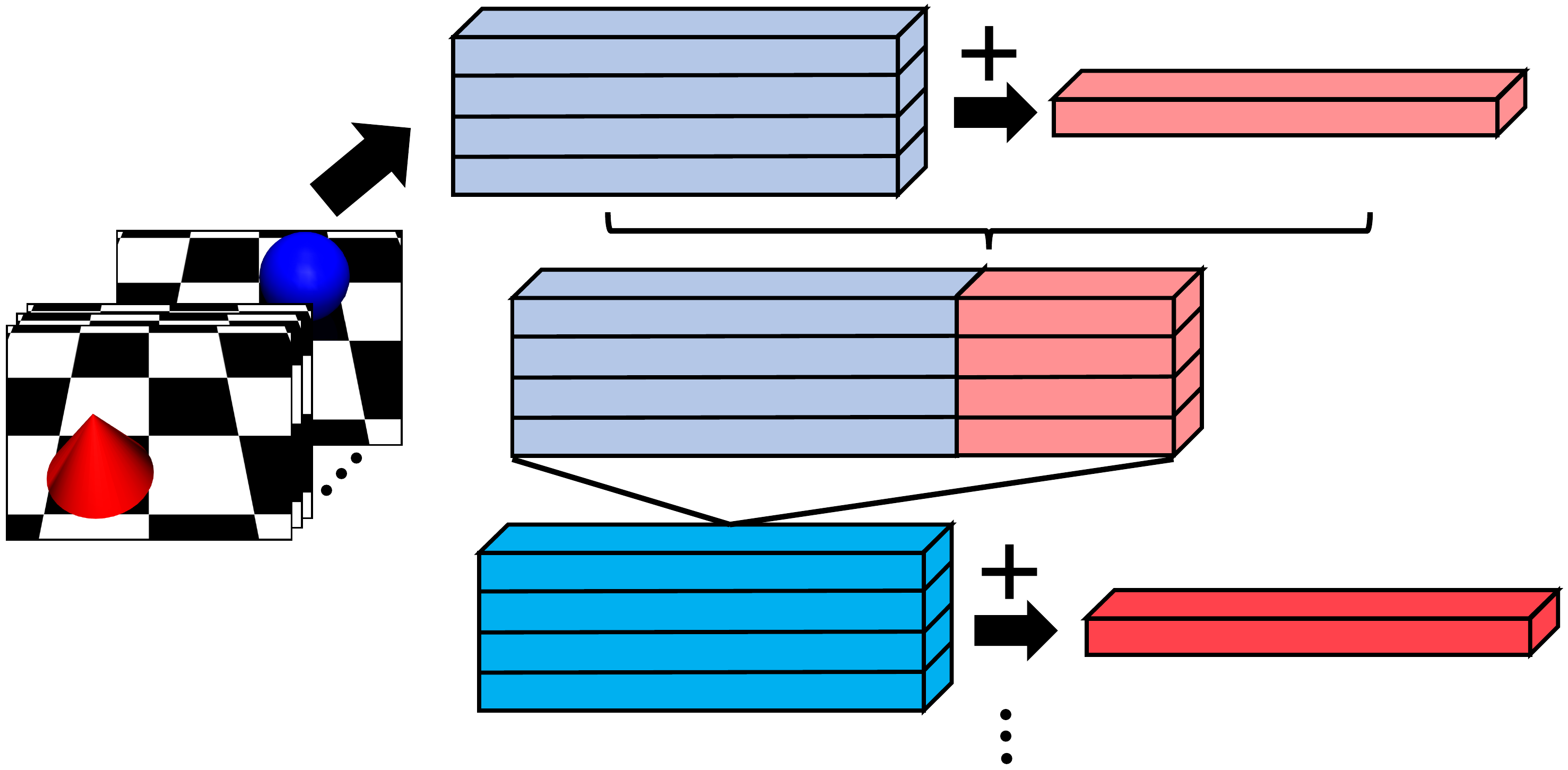}
    \caption{Candidates encoder.}
    \label{fig:struct}
\end{figure}

We align candidates embedding (multi-hot vectors or dense features) into a $1 \times |O| \times D$ tensor and apply $1 \times 1$ convolution to every candidate, where $D$ is the candidate embedding dimension. We sum the candidates embedding as the context embedding and concatenate it after each candidate's embedding, followed by another $1 \times 1$ convolution. This structure can be repeated as needed. Figure \ref{fig:struct} shows the structure. The final embedding of the candidates forms a tensor with shape $1 \times |O| \times D_M$, where $D_M$ is the dimension of the message encoding. Then we do another $1 \times 1 $ convolution with the message encoding of $m$ as the only tensor and get $|O|$ numbers, which is then fed into a sigmoid layer and returns the likelihood $P(m|o)$ for all $o \in O$. This likelihood then multiplies with the input prior and returns posterior $P(o|m)$ after normalization.
Teacher's Q-net reuses candidates last embedding layer. The $1 \times |O| \times D_M$ tensor is weighted by the return value of teacher's belief update function and teacher's ground truth belief (one-hot vector in referential games). We then sum the weighted tensors with respect to the second dimension and get two vectors. The concatenation of these vectors is passed into a fully connected layer and outputs a real number as the Q-value.

Student's policy network is relatively simple. We first pass in the output of student's belief update network to a fully connected layer followed by a sigmoid function, outputting the probability of waiting. Then action is sampled directly from the new belief. If wait, this action will be discarded, otherwise, it becomes the student's prediction of the target.
\end{appendices}
\end{document}